\documentclass[10pt,twocolumn]{article} 
\usepackage{simpleConference}
\usepackage{amsmath,graphicx}
\usepackage{xcolor}
\usepackage{amssymb}
\usepackage{algorithm}
\usepackage{algpseudocode}
\usepackage{stackengine}
\usepackage{tabularx,booktabs}
\usepackage{graphicx}
\usepackage{multirow}
\usepackage{hyperref}

\usepackage{xcolor}

\usepackage[nomargin,inline,draft]{fixme}
\fxsetup{theme=color,mode=multiuser}
\FXRegisterAuthor{pf}{apf}{\color{red}pascal}

\DeclareMathOperator*{\argmin}{argmin}
\makeatletter
\usepackage{forloop}
\newcounter{ct}
\newcommand{\markdent}[1]{\forloop{ct}{0}{\value{ct} < #1}{\hspace{-0.05cm}\hspace{\algorithmicindent}}}
\newcommand{\markcomment}[1]{\Statex\markdent{#1}}
\newcommand{\StateIndent}[1]{\State\markdent{#1}}

\begin{document}

\title{Block-Sparse Adversarial Attack to Fool Transformer-Based \\Text Classifiers}

\author{
Sahar Sadrizadeh\\{\textit{EPFL}}\\Lausanne, Switzerland\\ \href{mailto:sahar.sadrizadeh@epfl.ch}{ \tt\small{sahar.sadrizadeh@epfl.ch}}  \and 
Ljiljana Dolamic\\{\textit{Armasuisse S+T}}\\Thun, Switzerland\\\href{mailto:ljiljana.dolamic@armasuisse.ch}{ \tt\small{ljiljana.dolamic@armasuisse.ch}} \and
Pascal Frossard\\{\textit{EPFL}}\\Lausanne, Switzerland\\ \href{mailto:pascal.frossard@epfl.ch}{ \tt\small{pascal.frossard@epfl.ch}}
}

\maketitle
\thispagestyle{empty}

\begin{abstract}
Recently, it has been shown that, in spite of the significant performance of deep neural networks in different fields, those  are vulnerable to adversarial examples. In this paper, we propose a gradient-based adversarial attack against transformer-based text classifiers. The adversarial perturbation in our method is imposed to be block-sparse so that the resultant adversarial example differs from the original sentence in only a few words. Due to the discrete nature of textual data, we perform gradient projection to find the minimizer of our proposed optimization problem. Experimental results demonstrate that, while our adversarial attack  maintains the semantics of the sentence, it can reduce the accuracy of GPT-2 to less than 5\% on different datasets (AG News, MNLI, and Yelp Reviews). Furthermore, the block-sparsity constraint of the proposed optimization problem results in  small perturbations in the adversarial example. \footnote{ The source code of our attack can be found at \fontsize{7.5}{9}\selectfont \href{https://github.com/sssadrizadeh/transformer-text-classifier-attack}{https://github.com/sssadrizadeh/transformer-text-classifier-attack}}
\end{abstract}


\begin{figure*}[tb]
	\centering
	\includegraphics[page=1,width=.93\linewidth, trim={3.9cm 5.8cm 4.9cm 5cm},clip]{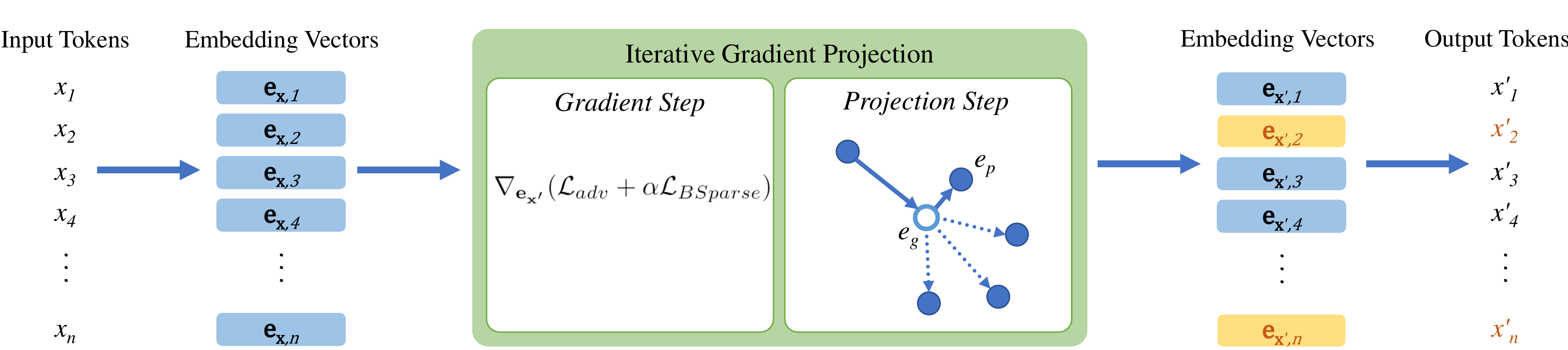}
	\caption{Block diagram of the proposed method.
	}
	\label{fig:blockdiagram}
\end{figure*}

\section{Introduction}

In recent years, with the emerging high computational devices, Deep Neural Networks (DNNs) have attracted tremendous attention in many different fields such as computer vision \cite{he2016deep} and Natural Language Processing (NLP) \cite{vaswani2017attention} due to their great performance. However, it has been shown that these models are highly vulnerable to perturbation of input samples, in particular to adversarial examples \cite{szegedy2014intriguing}. These examples, which are generated by making small or often imperceptible changes to the original input, can mislead the learning model to  classify the adversarial example into a wrong predetermined target class (targeted attack) or to a different class than the true one (untargeted attack). 
Recently, many methods have been proposed to generate adversarial examples in image data  to make the systems fail \cite{moosavi2016deepfool,madry2018towards}, but these methods cannot be directly extended to NLP models, due to both the different nature of the data representation and the difficulty of characterizing imperceptible changes in text.   

In visual applications, the main methods for generating adversarial examples are based on optimization and gradient descent. However, this is not readily extendable to textual data due to its specific nature. Therefore, there exists only a few white-box attacks, which have access to the parameters and gradients of the system, against NLP models. %\pfnote{next sentence not clear, rephrase} 
Although it is not possible to calculate the gradients in the discrete space of textual data, it has been proposed to find the gradients in the embedding space, which is continuous. For example, Papernot et al. \cite{papernot2016crafting} replace random words in the input sentence with the nearest word in the embedding space whose difference with the original word is in the direction of the gradient. Sato et al. \cite{sato2018interpretable} extends Adv-Text \cite{miyato2016adversarial} to generate adversarial perturbations by imposing the directions of the perturbations in the embedding space to align with meaningful embedding vectors. However, these methods may perturb many words in the sentence which makes the changes quite perceptible. On the other hand, Guo et al. \cite{guo2021gradient}, recently proposed Gradient-based Distributional Attack (GBDA) against text transformers.  They consider a probability distribution over all the vocabulary for each word in the adversarial sentence. They optimize this continuous matrix of distribution to fool the target model. However, their proposed formulation is highly over-parameterized.
%\pfnote{and? what is the main message? main problems with above methods?}

The second difficulty in dealing with textual data is the definition of imperceptibility of the adversarial attack. %  as it is not as straightforward as continuous data. 
The $\ell_p$-norm, which is common in images to measure the difference of adversarial example and the input, is not readily applicable in textual data.  There are different definitions for imperceptibility of adversarial attack in the literature. Some approximate it by the number of edits in the original text \cite{ebrahimi2018hotflip,gao2018black}. On the other hand, many attacks define imperceptibility as the semantic similarity between the adversarial example and the original input. They first select random words or find the most important words in the sentence based on different metrics such as the word saliency \cite{ren2019generating}. Afterwards, they replace the selected words with their synonyms \cite{ren2019generating,zang2020word}, other words with similar embedding vectors \cite{alzantot2018generating,jin2020bert}, or words predicted by a masked language model \cite{li2020bert}. However, most of these methods assume the black-box scenario and use heuristic strategies that result in sub-optimal performance. %There is also another type of attack, known as universal adversarial attack, which consists of a single snippet of text that can be added to any input sentence to generate an adversarial example \cite{behjati2019universal, wallace2019universal}. Thereby, the resultant adversarial example is perceptible to human eyes. %\pfnote{and? what is the main message of this paragraph? what argument does it support?}

%\pfnote{this paragraph is misplaced, it cuts the flow. It may actually not even be critical for the paper. } Adversarial attacks are also categorized into two groups considering the knowledge of the attacker of the parameters of the model:  white-box and black-box  attacks.   In  the  white-box  attack,  the  attacker  has  full knowledge  of  the learning model.  However, in the black-box scenario, the attacker only has access to the output of the DNN model for each input.  Although the black-box attack is more realistic since most of the time the model is unknown to the attacker, it has been shown that we can train a DNN model to distill the target model \cite{wallace2019universal,xiao2018generating}. 

%\pfnote{At this point, the reader should be clear on the objectives and limitations of related work. We might want to slightly revise the aove text accordingly.} 
In this paper, we propose a method based on gradient projection to generate token-level adversarial examples against transformer-based text classifiers. We assume the white-box scenario, which gives us access to the model parameters and also their gradients. We consider perturbing the sequence of embedding vectors of  the tokens in the input sentence. However, since we want only a few tokens to be changed, only a few blocks of the perturbation vector should be nonzero. Therefore, we add the block-sparsity constraint for the perturbation vector in the optimization problem. Moreover, we preserve the semantics of the sentence by projecting into the embedding vectors of the tokens which have the maximum cosine similarity with the embedding vectors of the corresponding original tokens. %Moreover, we use contextual embedding vectors of a language model with the same tokenizer as the target classifier in the imperceptibility term of the proposed optimization problem to maintain the semantic similarity to some extent. 
We evaluate our proposed attack against target transformer model with GPT-2 architecture \cite{radford2019language} fine-tuned for different downstream NLP tasks such as  natural language inference, sentiment analysis and news categorization. We compare our results with GDBA \cite{guo2021gradient}, a state-of-the-art white-box attack against text classifiers.  To our knowledge, GDBA is the only white-box attack in the literature against transformer models. Experimental results indicate that the proposed adversarial  attack achieves a competitive success rate in comparison to the GBDA method. Moreover, the projection to the closest token into the embedding space results in high semantic similarity between the adversarial example and the original sentence. Furthermore, our proposed block-sparsity constraint lead to small perturbations.  %\pfnote{and? what are the main results? And what are the main insights/perspectives/expected impact of the contributions?}

%\pfnote{The paper is not critical if we are short of space. If we keep it, we should stay at the level of Sections, and not go down to subsections probably. } 
The rest of this paper is organized as follows. In Section \ref{sec:problem}, we formulate the problem  of generating adversarial examples. Our attack algorithm is describe in Section \ref{sec:method}. We evaluate our algorithm against different transformer models and discuss the  experimental results in Section \ref{sec:experiment}. Finally, the paper is concluded in Section \ref{sec:conclusion}.

%\pfnote{it might be good to add a block scheme figure somewhere, for the general attack problem.}
\label{sec:intro}

\section{Problem Formulation}

In this section, we present the formulation of generating adversarial example for textual data in untargeted attacks. %\textcolor{red}{and targeted} attacks. 

Consider $f: \mathcal{X}\rightarrow \mathcal{Y}$ to be the target text classifier model which correctly predicts the class of the input sentence $\mathbf{x} \in \mathcal{X}$ to be $y = f(\mathbf{x}) \in \mathcal{Y}$. Every sentence is considered to be tokenized to a sequence of tokens. We are looking for an adversarial example $\mathbf{x'}$, which differs from the input sentence $\mathbf{x}$ in only a few tokens and is semantically similar to it. 
%\pfnote{token has not been defined}, fools the target model to predict wrongly, i.e., $f(\mathbf{x'}) \neq y$. 
However, the target model should classify  $\mathbf{x'}$ wrongly, i.e., $f(\mathbf{x'}) \neq y$. %However, to the human observer both $\mathbf{x}$ and $\mathbf{x'}$ should belong to the same class. 

Let $\mathbf{x} = x_1x_2...x_n$ be the input sentence which is a sequence of $n$ tokens of the vocabulary set $\mathcal{V}$. We assume that the adversarial example  $\mathbf{x'} = x'_1x'_2...x'_n$ is also a sequence of $n$ tokens. The tokens of these sentences are in a discrete space. Therefore, each of these tokens is transformed to a continuous vector, called an embedding vector, as the input of the target transformer model \cite{radford2019language}. %\pfnote{add reference and minimal short description for completeness}. 
Let $\text{emb}(.)$ denote the embedding function that gets a token as the intput and transforms it to a continuous vector. Therefore, we can represent the sentence $\mathbf{x}$  in the embedding space as a sequence of embedding vectors $\mathbf{e_x} = [\text{emb}(x_1),\text{emb}(x_2), ...,\text{emb}(x_n)]$ by transforming each of its tokens by the function $\text{emb}(.)$. %\pfnote{each token has its own embedding - please clarify and add reference}. 
Similarly, let $\mathbf{e_{x'}}= \mathbf{e_{x}} + \mathbf{r_{x}}$ represents the adversarial example as a sequence of embedding vectors. $\mathbf{r_x} = [\mathbf{r}_1,\mathbf{r}_2,...,\mathbf{r}_n]$ is the  sequence of the perturbation vectors of each token. % = \mathbf{e_x} - \mathbf{e_{x'}}

Now, in order to fool the model with an untargeted attack, we can find an adversarial example by maximizing the loss function of the classifier, i.e., cross entropy. This is equivalent to finding the perturbed sample $\mathbf{e_{x'}}$ that minimizes the loss $\mathcal{L}_{Adv}$, which is defined as the negative of the cross entropy: %\pfnote{to be complete, we might want to clarify the maximization versus minimization switch} %Moreover, we can use other adversarial loss such as the margin loss based on Carlini-Wagner (CW) attack \cite{carlini2017towards} as follows:
% \begin{equation}
%     \mathcal{L}_{adv} = - \mathcal{L}_f(\mathbf{e_x}+\mathbf{r},y),
% \end{equation}
\begin{equation}
    \mathcal{L}_{Adv} = - \mathcal{L}_f(\mathbf{e_{x'}},y), \label{eq.ladv}
\end{equation}
where $\mathcal{L}_f$ is the loss function of the model when the input is the adversarial example $\mathbf{e_{x'}}$ and the ground-truth class is $y$.

The above problem could lead to a large perturbation of the textual data. In order to constraint the changes to be small, we want to modify only a few tokens of the sentence. Therefore, only a few  perturbation vectors (some blocks of $\mathbf{r_x}$) that correspond to the modified tokens are non-zero, while others are zero. %This means that only a few blocks of the perturbation vector, corresponding to the perturbed tokens, are non-zero. 
In other words, the non-zero entries of the perturbation $\mathbf{r_x}$ occur in clusters, which means $\mathbf{r_x}$ should be block-sparse. 
%\pfnote{is it clear about why we talk about block-sparsity, and not mere sparsity? $r_i$'s are vectors.}. 
To impose the block-sparsity of the perturbation, we can impose the sparsity on the norm of each block \cite{elhamifar2012block}. %\pfnote{what is a block? is that sufficient to guarantee that the perturbation is small / 'imperceptible'?} 
Hence, in the final optimization problem, we will minimize the $\ell_1$ relaxation over all the $\ell_2$ norms of perturbation blocks $\mathbf{r}_i$ to ensure the sparsity of non-zero blocks: %\pfnote{this sentence is not clear}:
\begin{equation}
    \mathcal{L}_{BSparse} = \sum_{i=1}^n \|\mathbf{r}_i\|_2.
\end{equation}
%\pfnote{why 'emb' and not 'sparse' in L?}

%\pfnote{if you want to be consistent with introduction, where you argue for imperceptibility, you should probably motivate better the use of the sparse constraints, and how much it would lead to imperceptibility. This point might not be fully convincing ??}

% On the other hand, for the targeted attack, we can minimize the loss function for a specific target class ($t$), which is different from the ground-truth. 
% % \begin{equation}
% %     \mathcal{L}_{adv} = \mathcal{L}_f(\mathbf{e_x}+\mathbf{r},t).
% % \end{equation}
% \begin{equation}
%     \mathcal{L}_{adv} =  \mathcal{L}_f(\mathbf{e_{x'}},t),
% \end{equation}

Finally, we can reformulate the original optimisation problem of (\ref{eq.ladv}) by integrating the above block sparsity constraint. Therefore, our objective is to find the block-sparse perturbation that fool the target classifier by solving the following optimization problem:
% \begin{equation} \label{optimization}
%     \hat{\mathbf{r}} = \argmin_{\substack{\mathbf{r},\\\mathbf{e_x}+\mathbf{r}\in \mathcal{E_V}}} \mathcal{L}_{adv} + \alpha \mathcal{L}_{emb}, 
% \end{equation}
\begin{equation} \label{optimization}
    \hat{\mathbf{e_{x'}}} = \argmin_{\substack{\mathbf{e_{x'}}\in \mathcal{E_V}}} \mathcal{L}_{Adv} + \alpha \mathcal{L}_{BSparse}, 
\end{equation}
where $\mathcal{E_V}$ is the discrete subspace of every token %\pfnote{what is 'every vocabulary'? there are several of them? clarify} 
of the vocabulary set $\mathcal{V}$ in the embedding space. Moreover, $\alpha$ is the hyper-parameter that determines the relative importance of the block-sparsity term. %We propose an algorithm to solve this problem in the next Section. 

\label{sec:problem}

\section{Proposed Method}

%Since we are dealing with textual data, the proposed optimization problem is discrete and we will use gradient projection to solve it.  Hence, we consider the optimization problem to be continuous, and every few iterations, we project the result to the discrete space.  As illustrated in Figure \ref{fig:blockdiagram}, we first transform the tokens of the input sentence to the continuous vectors in the embedding space. Algorithm \ref{alg} presents the pseudo-code of the proposed method for finding the minimizer of the discrete optimization problem \eqref{optimization}.
% \setlength{\textfloatsep}{20pt}
\begin{algorithm}[!tb]
\small
\caption{Block-Sparse Adversarial Attack}
\label{alg}
\begin{algorithmic}[1]
\State \textbf{Input}:
\markcomment{1} $f(.)$: Target classifier model, $\mathcal{V}$: Vocabulary set
\markcomment{1} $\mathbf{x}$ : Tokenized input sentence, $lr$: Learning rate
% \markcomment{1} 
\markcomment{1} $\mathrm{A}$: Set of decreasing values for Hyper-parameter $\alpha$ to \markcomment{1} control the importance of the block-sparsity term
% \markcomment{1} 
\markcomment{1} $K$: Maximum number of iterations
\State \textbf{Output}:
\markcomment{1} $\mathbf{x'}$: Generated adversarial example
\Procedure{}{}
\markcomment{1} \textbf{initialization:}
\StateIndent{1} $\textbf{buffer}\leftarrow\text{empty}$, $y\leftarrow f(\mathbf{x})$, $k \leftarrow 0$
% \StateIndent{1} 
\StateIndent{1} $\forall i \in \{1,...n\} \quad \mathbf{e}_{\mathbf{g},i}\leftarrow \text{emb}(x_i)$ 
\For {$\alpha$ in $\mathrm{A}$}
\While {$f(\mathbf{e}_{\mathbf{p}})=y \;\text{and} \;k \le K$}
\State $k \leftarrow k + 1$
\markcomment{3} \textbf{Step 1:} Gradient descent in the continuous 
\markcomment{3} embedding space:
\StateIndent{1} $\mathbf{e_g} \leftarrow \mathbf{e}_{\mathbf{g}} - lr. \nabla_{\mathbf{e_{x'}}} (\mathcal{L}_{adv} + \alpha \mathcal{L}_{BSparse})$
\markcomment{3} \textbf{Step 2:} Projection  to the discrete subspace $\mathcal{E_V}$ 
\markcomment{3} and update if the sentence is new: 
\StateIndent{1} \textbf{for} {$i \in \{1,...,n\}$} \textbf{do}
\StateIndent{2} $\mathbf{e}_{\mathbf{p},i} \leftarrow \argmin\limits_{\mathbf{e} \in \mathcal{E_V}} \frac{\mathbf{e}^\top \mathbf{e}_{\mathbf{g},i}}{\|\mathbf{e}\|_2.\|\mathbf{e}_{\mathbf{g},i}\|_2}$
\StateIndent{1} \textbf{end for}
% \markcomment{2} \textbf{Step 3:} Update with the projection step if the
% \markcomment{2}  sentence is new:
\StateIndent{1} \textbf{if} {$\mathbf{e_p}$ not in \textbf{buffer}} \textbf{then}
\StateIndent{2} add $\mathbf{e_p}$ to \textbf{buffer}
\StateIndent{2} $\mathbf{e_g}\leftarrow \mathbf{e}_{\mathbf{p}}$
\StateIndent{1} \textbf{end if}

\EndWhile
\If {$f(\mathbf{e}_{\mathbf{p}})\neq y$}
\State break (adversarial example is found)
\EndIf
\EndFor
\State\Return ${\mathbf{e_{x'}}} \leftarrow \mathbf{e_p}$
\EndProcedure
\end{algorithmic}
\normalsize

\end{algorithm}

In this section, we  explain our algorithm to find the solution of the proposed optimization problem \eqref{optimization}. The block diagram of our method can be found in Figure \ref{fig:blockdiagram}. As depicted in this figure, we first transform each token of the input sentence to a continuous embedding vector and then we use gradient projection to solve the optimization problem \eqref{optimization}. 

\begin{table*}[!t]
	\centering
		\renewcommand{\arraystretch}{.85}

	\setlength{\tabcolsep}{3pt}
	\caption{Examples of successful adversarial examples on different datasets.}
	\scalebox{0.75}{
	   % \fontsize{11}{12}\selectfont
		\begin{tabular}[t]{@{} l|l|l| >{\parfillskip=0pt}p{16.5cm} @{}}
			\toprule[1pt]
		    \textbf{Dataset} & \textbf{Sentence}  & \textbf{Prediction} & \textbf{Text}\\
		
			\midrule[1pt]
			
% 			\midrule[1pt]
			\multirow{6}{*}{\textbf{MNLI}}  & \multirow{3}{*}{Original} &  \multirow{3}{*}{Neutral (97.26\%)} & Premise: In the summer, the Sultan's Pool, a vast outdoor amphitheatre, stages rock concerts\\
			
			& & & or other big-name events. \hfill\mbox{}\\
			
			& & & Hypothesis: \textcolor{blue}{\textbf{Most}} rock concerts take place in the Sultan's Pool amphitheatre.\hfill\mbox{} \\
			
			\cline{2-4}
			\rule{0pt}{2.5ex} 
			
			 & \multirow{3}{*}{Adversarial} &  \multirow{3}{*}{\textcolor{red}{Entailment (99.19\%)}}  & Premise: In the summer, the Sultan's Pool, a vast outdoor amphitheatre, stages rock concerts\\
			 & & & or other big-name events.\hfill\mbox{}\\
			 
			 & & & Hypothesis: \textcolor{red}{\textbf{Many}} rock concerts take place in the Sultan's Pool amphitheatre.\hfill\mbox{} \\
			\midrule[1pt]
			
			\multirow{4}{*}{\textbf{AG News}} & \multirow{2}{*}{Original} &  \multirow{2}{*}{Sci/Tech (99.39\%)} & Motorola and HP in \textcolor{blue}{\textbf{Linux}} tie-up Motorola plans to sell mobile phone network equipment that uses \\
			
			& & & \textcolor{blue}{\textbf{Linux}}-based code, a step forward in network gear makers \#39; efforts to rally around a standard.\hfill\mbox{}  \\
			 
			\cline{2-4}
			\rule{0pt}{2.5ex} 
			 
			& \multirow{2}{*}{Adversarial} &  \multirow{2}{*}{\textcolor{red}{Business (83.56\%)}} & Motorola and HP in \textcolor{red}{\textbf{PC}} tie-up Motorola plans to sell mobile phone network equipment that  uses\\ 
			
			& & & \textcolor{red}{\textbf{PC}}-based code. a step forward in network gear makers \#39; efforts to rally around a standard.\hfill\mbox{} \\
			\midrule[1pt]
			
			\multirow{6}{*}{\textbf{Yelp}} & \multirow{3}{*}{Original} & \multirow{3}{*}{Negative (99.90\%)} & This place holds a nostalgic appeal for people born and raised in Pittsburgh who grew up eating\\
			& & & here. If that experience is what your looking for, please visit. If you're looking for a tasty meal,\\
			& & & go somewhere else. 5 stars for history, \textcolor{blue}{\textbf{0}} for food quality and flavor.\hfill\mbox{} \\
			
			\cline{2-4}
			\rule{0pt}{2.5ex} 
			
			 & \multirow{3}{*}{Adversarial} & \multirow{3}{*}{\textcolor{red}{ Positive (96.54\%)}} & This place holds a nostalgic appeal for people born and raised in Pittsburgh who grew up eating\\
			 & & & here. If that experience is what your looking for, please visit. If you're looking for a tasty meal,\\
			 & & & go somewhere else. 5 stars for history,\textcolor{red}{\textbf{1}} for food quality and flavor.\hfill\mbox{} \\
			
			\bottomrule[1pt]
		\end{tabular}
	}
	\label{tab:sample}
\end{table*}

Since we are dealing with textual data, \eqref{optimization} is a discrete optimization problem. In other words, the tokens of the resultant adversarial example should be in the vocabulary set $\mathcal{V}$; hence $\mathbf{e_{x'}}$ should be in the discrete subspace $\mathcal{E_V}$. First, we consider $\mathbf{e_{x'}}$ to be in the embedding space  $\mathcal{E}$ (and not necessarily in $\mathcal{E_V}$). Thus, we can perform gradient descent to solve the optimization problem \eqref{optimization}. In each iteration of our algorithm, we first update %the perturbation vectors regarding all the tokens of the adversarial example
the embedding vectors of all the tokens of the adversarial example in the continuous space $\mathcal{E}$. Let $\mathbf{e}_{\mathbf{g},i}$ denote this updated vector in the continuous space corresponding to the $i$-th token. Afterwards, we project the updated embedding vectors $\mathbf{e}_{\mathbf{g},i}$, which may not necessarily correspond to a token in the vocabulary $\mathcal{V}$, to the embedding vectors of the closest meaningful tokens. We use cosine similarity metric to find the closest embedding vectors in $\mathcal{E_V}$ and apply the projection for each token independently: %\pfnote{this paragraph should be rephrased - it goes rapidly into technical details - it might be better to give the general ideas on how to solve the optimisation problem first, what techniques are used, the general organisation of the algorithm, and then go in the relevant details. }
\begin{equation}
\forall i \in \{1,...,n\}: \quad
    \mathbf{e}_{\mathbf{p},i} = \argmin_{\mathbf{e} \in \mathcal{E_V}} \frac{\mathbf{e}^\top \mathbf{e}_{\mathbf{g},i}}{\|\mathbf{e}\|_2.\|\mathbf{e}_{\mathbf{g},i}\|_2}.
\end{equation}

Furthermore, since we are dealing with discrete data, it is possible that through iterations we come across a previously computed embedding vector after the projection. Moreover, if the perturbation vector is too small, the updated vectors will be projected to the previous sentence. In these cases, the algorithm will be stuck in a loop as the computed gradients will stay the same. To prevent such undesirable scenarios, we  update the embedding vectors by the projection step only when the projected sentence has not been generated before. To this end, we save all the updated sentences in a buffer, and  update the embedding vectors by the projected ones only if the output of the projection step is not in the buffer. %\pfnote{it appears a bit technical, a bit adhoc} 
%Therefore, the algorithm may go through several gradient descent step before a projection step. 
%line here removed at last moment 
These steps are performed iteratively until the target model is fooled or a maximum number of iterations is reached (the algorithm fails to find an adversarial example in this case). 

As another consideration, we do not fix the value of $\alpha$ in \eqref{optimization}, which determines the importance of imperceptibility. For higher values of $\alpha$, the algorithm tries to find adversarial examples with smaller token error rate that are more similar to the original sentence. However, by increasing $\alpha$, the success rate of finding an adversarial example decreases. Therefore, we will consider a large value for this hyper-parameter at first. If the algorithm fails  to find an adversarial example, we will decrease the value of $\alpha$.
Algorithm \ref{alg} presents the pseudo-code of the proposed method for finding the minimizer of the discrete optimization problem \eqref{optimization}.

% \stackunder{\text{argmin}}
\label{sec:method}

 \section{Experimental Result}

In this section, we evaluate our proposed adversarial attack in different text classification tasks such as natural language inference, sentiment classification, and news categorization. %The source codes of our experiments are publicly available to allow reproducing our results\footnote{The source code will be available upon acceptance of the paper.}.

\vspace{-5pt}
\paragraph*{Datasets.} We evaluate our proposed method on the test set of three datasets: 
% \begin{itemize}
%     \item \textbf{MNLI} 
%     \item \textbf{AG News}
%     \item \textbf{Yelp Reviews}
% \end{itemize}
MNLI\footnote{We evaluate our method on the matched validation set of  MNLI dataset.} \cite{williams2018broad} (natural language inference), AG News \cite{zhang2015character} (news categorization), and Yelp Reviews \cite{zhang2015character} (sentiment classification). Some statistics of these datasets can be found in Table \ref{tab:dataset}. 

\vspace{-5pt}
\paragraph*{Model.} For the target model, we fine-tuned a pre-trained transformer model with GPT-2 architecture on all the mentioned datasets. 

\vspace{-5pt}
\paragraph*{Baseline.} We compare the result of our method with that of GDBA \cite{guo2021gradient}. To our knowledge, GDBA is the only white-box attack in the literature against transformer models.  %in terms of success rate (accuracy of the target model after attack) and semantic similarity

\vspace{-5pt}
\paragraph*{Hyper-parameters.} We use Adam optimizer to find the minimizer of the proposed optimization problem with learning rate $\{0.15,0.3\}$. Moreover, the coefficient $\alpha$ in \eqref{optimization} is in the set $\{10,8,5,2\}$ divided by the length of the sentence. For larger values of learning rate and smaller values of $\alpha$, the attack is more aggressive and more words of the sentence are modified. Therefore we will change them in the mentioned sets if only the attack fails.   

\vspace{-5pt}
\paragraph*{Evaluation.} 
To evaluate our adversarial attack, we solve \eqref{optimization} by algorithm \ref{alg} and attack the target model with GPT-2 architecture which is fine-tuned over one of the aforementioned datasets. The MNLI dataset consists of sentence pairs, premise and hypothesis, and we attack them separately.  Table \ref{tab:result} shows the result of our method in comparison with GDBA in terms of accuracy of the target model after the adversarial attack and semantic similarity. Semantic similarity between the adversarial example and the original sentence  is computed by Universal Sentence Encoders \cite{cer2018universal} as is common in the literature. It is worth mentioning that we consider that our attack has failed if the semantic similarity is less than a threshold. 
The results show that our attack successfully drop the accuracy of the target model to less than 5\% for all of the datasets while the semantic similarity is preserved (more than 0.8 in all cases). Compared to GBDA, success rate of our attack is superior in all cases, except for the MNLI dataset, in which our method achieves competitive results. %ours is only 0.2\% less. %As shown in Table \ref{tab:result}, the performance of our attack against MNLI dataset is slightly lower
Table \ref{tab:sample} also shows some adversarial examples against different datasets generated by our method.

\setlength{\textfloatsep}{20pt}
\begin{table}[tbp]
	\centering
		\renewcommand{\arraystretch}{0.8}
	\setlength{\tabcolsep}{3pt}
	\caption{Some statistics of the evaluation datasets. Last column (Clean Acc.(\%)) is the accuracy of the fine-tuned GPT-2.}
	\label{tab:dataset}
	\scalebox{.85}{
		\begin{tabular}[t]{@{} lcccc @{}}
			\toprule[1pt]
		    \textbf{Dataset}  & {\textbf{Avg. Length}} &   {\textbf{\#Classes}} &   {\textbf{Test Set}} &  {\textbf{Clean Acc.}}\\
% 			\hline
% 			\rule{0pt}{2.5ex}  
			\midrule[1pt]
			Ag News &  43 & 4 & 7600 & 94.8 \\
			%& & \textbf{(LFW)}  & \textbf{(MOBIO)}\\
			\midrule
			MNLI &  11 & 3 & 9815 & 81.7  \\ \midrule%\cline{2-7} \rule{0pt}{2.5ex}
			Yelp Reviews&  157 & 2 &  38000 & 97.8 \\
			
			\bottomrule[1pt]
		\end{tabular}
	}
\end{table}

\setlength{\textfloatsep}{20pt}
\begin{table}[tbp]
	\centering
		\renewcommand{\arraystretch}{0.8}
	\setlength{\tabcolsep}{3pt}
	\caption{Performance of white-box attack against the fine-tuned GPT-2 in terms of  after attack accuracy (Adv. Acc.(\%)) and semantic similarity (Sim.). For the MNLI dataset, the results of attacking the hypothesis sentences are in brackets.}
	\label{tab:result}
	\scalebox{0.8}{
		\begin{tabular}[t]{@{} lcccccccc @{}}
			\toprule[1pt]
		    \multirow{2}{*}{\textbf{Method}}  & \multicolumn{2}{c}{\textbf{Ag News}} &&   \multicolumn{2}{c}{\textbf{MNLI}} &&   \multicolumn{2}{c}{\textbf{Yelp Reviews}}\\
			\cline{2-3}
			\cline{5-6}
			\cline{8-9}
			\rule{0pt}{2.5ex}    
			&  {Adv. Acc.} & {Sim.} && {Adv. Acc.} & {Sim.} &&  {Adv. Acc.} & {Sim.} \\
			%& & \textbf{(LFW)}  & \textbf{(MOBIO)}\\
			\midrule[1pt]
			Proposed &  0.4 & 0.87 && 3.0 (1.3) & 0.85 (0.82) && 1.8 & 0.87 \\
% 			\midrule[1pt]
% 			Proposed &  0.8 & 0.87 && 4.9 & 0.86 && 2.5 & 0.88 \\
% 			\midrule[1pt]
% 			Proposed &  1.6 & 0.88 && 7.7 & 0.87 && 3.7 & 0.88 \\
			\midrule%\cline{2-7} \rule{0pt}{2.5ex}
			GBDA &  6.6 & 0.90 &&  2.8 (11.0) & 0.82 (0.88) && 2.9 & 0.94\\
			
			\bottomrule[1pt]
		\end{tabular}
	}
\end{table}

\setlength{\textfloatsep}{20pt}
\begin{figure}[t]
	\centerline{
		\includegraphics[width=.95\linewidth]{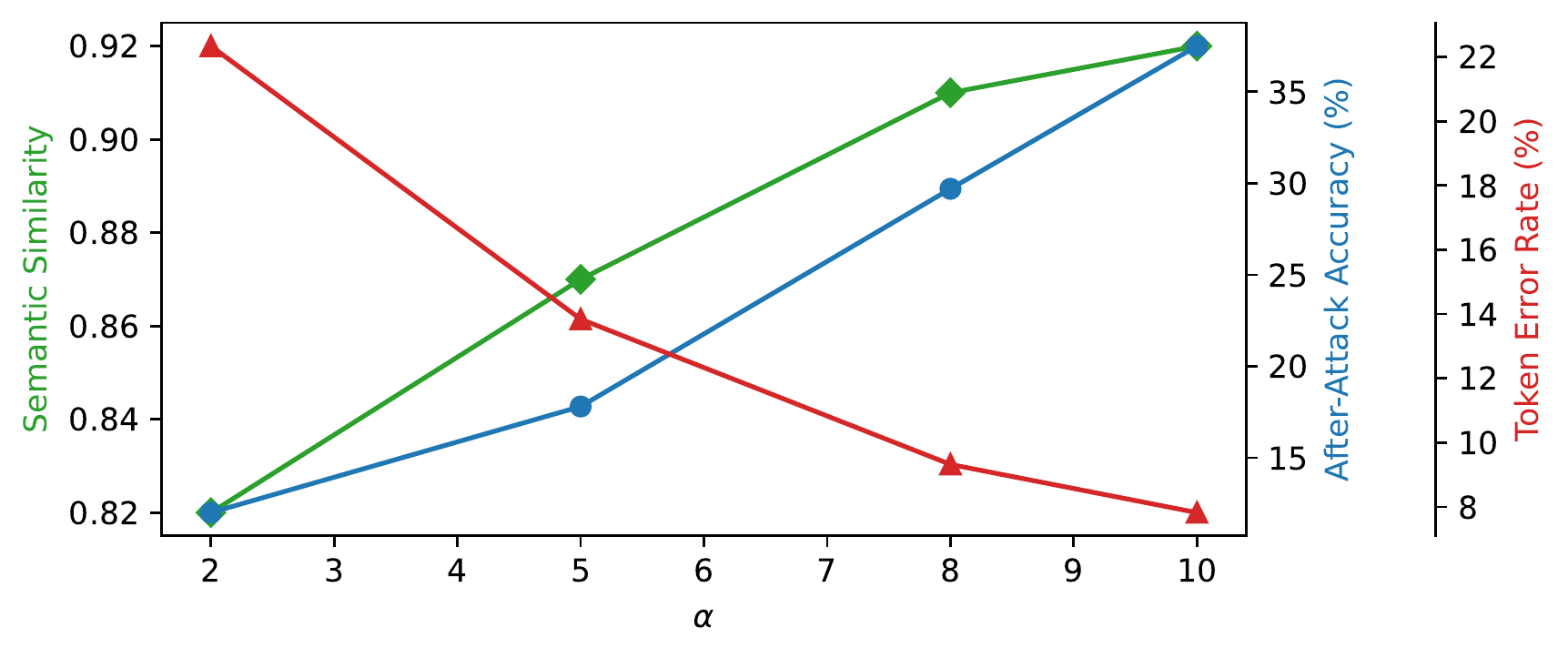}}
	 \vspace{-10pt}\caption{Effect of the hyper-parameter $\alpha$ on the performance.} %of our attack.}
	\label{fig:alpha}
\end{figure}

We investigate the effect of the hyper-parameter $\alpha$ in the  optimization problem \eqref{optimization} on the performance of our method on AG News dataset. Figure \ref{fig:alpha} depicts the effect of this hyper-parameter on the accuracy of the target model, semantic similarity, and token error rate. By increasing $\alpha$, success rate of our attack decreases while semantic similarity increases and the token error rate decreases. It is worth mentioning that we fix the learning at 0.15 for this experiment. Therefore, the accuracy is lower than the one reported in Table \ref{tab:result}.

\label{sec:experiment}

\section{Conclusion}

In this paper, we proposed a new white-box  attack based on gradient projection against text classifiers. We proposed an optimization problem with a block-sparsity constraint to ensure that only a few words of the sentence are modified. Experimental results show that our attack is highly effective on fooling text classifiers in different tasks and it preserves the semantics of the sentence. In all tasks,  the accuracy of the target model drops to less than 5\% and the semantic similarity is more than 80\%. We also compared our attack with GDBA, the only white-box attack against transformers. The success rate of our attack is superior to GDBA in all cases except for the MNLI dataset, in which our method achieves comparable results to GDBA.

\label{sec:conclusion}

\small
\bibliographystyle{abbrv}
\bibliography{refs}

\begin{thebibliography}{10}

\bibitem{alzantot2018generating}
M.~Alzantot, Y.~Sharma, A.~Elgohary, B.-J. Ho, M.~Srivastava, and K.-W. Chang.
\newblock Generating natural language adversarial examples.
\newblock In {\em Proceedings of the 2018 Conference on Empirical Methods in
  Natural Language Processing}, pages 2890--2896, 2018.

\bibitem{cer2018universal}
D.~Cer, Y.~Yang, S.-y. Kong, N.~Hua, N.~Limtiaco, R.~S. John, N.~Constant,
  M.~Guajardo-C{\'e}spedes, S.~Yuan, C.~Tar, et~al.
\newblock Universal sentence encoder.
\newblock {\em arXiv preprint arXiv:1803.11175}, 2018.

\bibitem{ebrahimi2018hotflip}
J.~Ebrahimi, A.~Rao, D.~Lowd, and D.~Dou.
\newblock Hotflip: White-box adversarial examples for text classification.
\newblock In {\em Proceedings of the 56th Annual Meeting of the Association for
  Computational Linguistics (Volume 2: Short Papers)}, pages 31--36, 2018.

\bibitem{elhamifar2012block}
E.~Elhamifar and R.~Vidal.
\newblock Block-sparse recovery via convex optimization.
\newblock {\em IEEE Transactions on Signal Processing}, 60(8):4094--4107, 2012.

\bibitem{gao2018black}
J.~Gao, J.~Lanchantin, M.~L. Soffa, and Y.~Qi.
\newblock Black-box generation of adversarial text sequences to evade deep
  learning classifiers.
\newblock In {\em 2018 IEEE Security and Privacy Workshops (SPW)}, pages
  50--56. IEEE, 2018.

\bibitem{guo2021gradient}
C.~Guo, A.~Sablayrolles, H.~J{\'e}gou, and D.~Kiela.
\newblock Gradient-based adversarial attacks against text transformers.
\newblock {\em arXiv preprint arXiv:2104.13733}, 2021.

\bibitem{he2016deep}
K.~He, X.~Zhang, S.~Ren, and J.~Sun.
\newblock Deep residual learning for image recognition.
\newblock In {\em Proceedings of the IEEE conference on computer vision and
  pattern recognition}, pages 770--778, 2016.

\bibitem{jin2020bert}
D.~Jin, Z.~Jin, J.~T. Zhou, and P.~Szolovits.
\newblock Is bert really robust? a strong baseline for natural language attack
  on text classification and entailment.
\newblock In {\em Proceedings of the AAAI conference on artificial
  intelligence}, volume~34, pages 8018--8025, 2020.

\bibitem{li2020bert}
L.~Li, R.~Ma, Q.~Guo, X.~Xue, and X.~Qiu.
\newblock Bert-attack: Adversarial attack against bert using bert.
\newblock In {\em Proceedings of the 2020 Conference on Empirical Methods in
  Natural Language Processing (EMNLP)}, pages 6193--6202, 2020.

\bibitem{madry2018towards}
A.~Madry, A.~Makelov, L.~Schmidt, D.~Tsipras, and A.~Vladu.
\newblock Towards deep learning models resistant to adversarial attacks.
\newblock In {\em International Conference on Learning Representations, ICLR
  2018}, 2018.

\bibitem{miyato2016adversarial}
T.~Miyato, A.~M. Dai, and I.~Goodfellow.
\newblock Adversarial training methods for semi-supervised text classification.
\newblock {\em arXiv preprint arXiv:1605.07725}, 2016.

\bibitem{moosavi2016deepfool}
S.-M. Moosavi-Dezfooli, A.~Fawzi, and P.~Frossard.
\newblock Deepfool: a simple and accurate method to fool deep neural networks.
\newblock In {\em Proceedings of the IEEE conference on computer vision and
  pattern recognition}, pages 2574--2582, 2016.

\bibitem{papernot2016crafting}
N.~Papernot, P.~McDaniel, A.~Swami, and R.~Harang.
\newblock Crafting adversarial input sequences for recurrent neural networks.
\newblock In {\em MILCOM 2016-2016 IEEE Military Communications Conference},
  pages 49--54. IEEE, 2016.

\bibitem{radford2019language}
A.~Radford, J.~Wu, R.~Child, D.~Luan, D.~Amodei, I.~Sutskever, et~al.
\newblock Language models are unsupervised multitask learners.
\newblock {\em OpenAI blog}, 1(8):9, 2019.

\bibitem{ren2019generating}
S.~Ren, Y.~Deng, K.~He, and W.~Che.
\newblock Generating natural language adversarial examples through probability
  weighted word saliency.
\newblock In {\em Proceedings of the 57th annual meeting of the association for
  computational linguistics}, pages 1085--1097, 2019.

\bibitem{sato2018interpretable}
M.~Sato, J.~Suzuki, H.~Shindo, and Y.~Matsumoto.
\newblock Interpretable adversarial perturbation in input embedding space for
  text.
\newblock In {\em Proceedings of the 27th International Joint Conference on
  Artificial Intelligence}, pages 4323--4330, 2018.

\bibitem{szegedy2014intriguing}
C.~Szegedy, W.~Zaremba, I.~Sutskever, J.~Bruna, D.~Erhan, I.~Goodfellow, and
  R.~Fergus.
\newblock Intriguing properties of neural networks.
\newblock In {\em 2nd International Conference on Learning Representations,
  ICLR 2014}, 2014.

\bibitem{vaswani2017attention}
A.~Vaswani, N.~Shazeer, N.~Parmar, J.~Uszkoreit, L.~Jones, A.~N. Gomez,
  {\L}.~Kaiser, and I.~Polosukhin.
\newblock Attention is all you need.
\newblock In {\em Advances in neural information processing systems}, pages
  5998--6008, 2017.

\bibitem{williams2018broad}
A.~Williams, N.~Nangia, and S.~Bowman.
\newblock A broad-coverage challenge corpus for sentence understanding through
  inference.
\newblock In {\em Proceedings of the 2018 Conference of the North American
  Chapter of the Association for Computational Linguistics: Human Language
  Technologies, Volume 1 (Long Papers)}, pages 1112--1122, 2018.

\bibitem{zang2020word}
Y.~Zang, F.~Qi, C.~Yang, Z.~Liu, M.~Zhang, Q.~Liu, and M.~Sun.
\newblock Word-level textual adversarial attacking as combinatorial
  optimization.
\newblock In {\em Proceedings of the 58th Annual Meeting of the Association for
  Computational Linguistics}, pages 6066--6080, 2020.

\bibitem{zhang2015character}
X.~Zhang, J.~Zhao, and Y.~LeCun.
\newblock Character-level convolutional networks for text classification.
\newblock {\em Advances in neural information processing systems}, 28:649--657,
  2015.

\end{thebibliography}

\end{document}